\pdfoutput=1
\PassOptionsToPackage{dvipsnames}{xcolor}
\documentclass[11pt, dvipsnames]{article}

\usepackage[]{acl}

\usepackage{times}
\usepackage{latexsym}
\usepackage{array}
\usepackage{booktabs}
\usepackage{siunitx}
\usepackage{amsmath,amssymb,amsthm}
\usepackage{multirow}
\usepackage[ruled,vlined]{algorithm2e}
\usepackage{ragged2e}

\usepackage[T1]{fontenc}

\usepackage[utf8]{inputenc}

\usepackage{microtype}
\usepackage{subfig}
\usepackage{tabularx}
\usepackage{enumitem}

\SetArgSty{textnormal}
\usepackage{graphicx}

\usepackage{colortbl}         
\usepackage{amssymb}       
\usepackage[dvipsnames]{xcolor}
\usepackage{marvosym}

\title{Subword Segmental Machine Translation: \\Unifying Segmentation and Target Sentence Generation}
\author{Francois Meyer and Jan Buys\\
  Department of Computer Science \\
  University of Cape Town \\
  \texttt{MYRFRA008@myuct.ac.za, jbuys@cs.uct.ac.za}}
\begin{document}
\maketitle
\begin{abstract}
Subword segmenters like BPE operate as a preprocessing step in neural machine translation and other (conditional) language models.
They are applied to datasets before training, so translation or text generation quality relies on the quality of segmentations. We propose a departure from this paradigm, called subword segmental machine translation (SSMT). 
SSMT unifies subword segmentation and MT in a single trainable model. It learns to segment target sentence words while jointly learning to generate target sentences. 
To use SSMT during inference we propose  \emph{dynamic decoding}, a text generation algorithm that adapts segmentations as it generates translations.
Experiments across 6 translation directions show that SSMT improves chrF scores for morphologically rich agglutinative languages. Gains are strongest in the very low-resource scenario.
SSMT also learns subwords that are closer to morphemes compared to baselines and proves more robust on a test set constructed for evaluating morphological compositional generalisation.

% that that evaluate how well models handle novel combinations of known morphemes. 
%SSMT outperforms baseline MT models 
%SSMT again outperforms standard subword-based models, showing that it learns how to compose morphemes into words.
\end{abstract}

\section{Introduction}

The continued success of neural machine translation (NMT) can be partially attributed to effective subword segmenters. Algorithms like byte-pair encoding (BPE) \citep{sennrich-etal-2016-neural} and Unigram LM (ULM) \citep{kudo-2018-subword} are computationally efficient preprocessing steps that enable smaller vocabularies and open-vocabulary models. 

These methods have proved quite effective, but fall short in certain contexts. For morphologically complex languages they are sub-optimal \citep{klein-tsarfaty-2020-getting} and inconsistent \citep{meyer-buys-2022-subword}. 
%, not universally applicable across languages and tasks
%\citep{zhu-etal-2019-systematic}, and over-segment words for low-resource languages  .
This is amplified in low-resource settings \citep{zhu-etal-2019-importance, wang-etal-2021-multi-view, acs-2021-exploring}, where handling rare words is crucial. 
These issues can be partially attributed to the fact that subword segmentation is separated from model training.
BPE and ULM are applied to the training corpus before training starts, so models are reliant on their output.
This is not ideal, since these algorithms do not learn segmentations that optimise model performance. %They are language-agnostic and task-independent.

%they are in no way connected to the model being trained or the task being trained for. 

\citet{he-etal-2020-dynamic} address this issue by proposing dynamic programming encoding (DPE), which trains an NMT model that marginalises over target sentence segmentations. After training they apply their model as a subword segmenter by computing the maximising segmentations.
%, and train new NMT models on the resulting segmented corpora. 
DPE is still a preprocessing step (a separate vanilla NMT model is trained on a corpus segmented by DPE), but since its segmentations are trained on MT, they are at least connected to the task at hand.

\begin{figure}[t]
    \centering
	\begin{tabular}{lll} 
		\toprule
	     %\multicolumn{2}{c}{\textbf{sesihambe}}\\
	     %\midrule
          \textbf{Train} & &\\
            \midrule
	     \textcolor{Blue}{I} \textcolor{Red}{do} \textcolor{Green}{understand}. & $\rightarrow $&   \textcolor{Blue}{Ndi}-\textcolor{Red}{ya}-\textcolor{Green}{qonda}. \\
	      \textcolor{Blue}{I} am tired. & $\rightarrow $&   \textcolor{Blue}{Ndi}-diniwe. \\
	     Where are \textcolor{Purple}{you} from? & $\rightarrow $&  \textcolor{Purple}{U}-vela phi?\\
         Are \textcolor{Purple}{you} \textcolor{Brown}{busy}?  & $\rightarrow $ & Ingaba \textcolor{Purple}{u}-\textcolor{Brown}{xakekile}? \\
        \midrule
          \textbf{Test} & & \\
           \midrule
	     \textcolor{red}{Do} \textcolor{Purple}{you} \textcolor{Green}{understand}? & $\rightarrow $& \textcolor{Purple}{U}-\textcolor{Red}{ya}-\textcolor{Green}{qonda}?\\
	      \textcolor{Blue}{I} am \textcolor{Brown}{busy}.  & $\rightarrow $&   \textcolor{Blue}{Ndi}-\textcolor{Brown}{xakekile}.\\
		\bottomrule
	\end{tabular}
	\captionof{table}{Parallel English-Xhosa sentences with morphologically segmented Xhosa words. The train/test split shows why its critical to accurately model morphemes and morphological compositional generalisation i.e. novel combinations of known morphemes.} 	\label{examples}
	%\vspace{-2cm}
\end{figure}

In this paper we go one step further by fully unifying NMT and subword segmentation. We propose  subword segmental machine translation (SSMT), an end-to-end NMT model that learns subword segmentation during training and can be used directly for inference. It is trained with a dynamic programming algorithm that enables  learning subword segmentations that optimise its MT training objective. The architecture is a Transformer-based adaptation of the subword segmental language model (SSLM) \citep{meyer-buys-2022-subword} for the joint task of MT and target-side segmentation.

We also propose \emph{dynamic decoding}, a decoding algorithm for subword segmental models that dynamically adapts subword segmentations as it generates translations. 
The fact that our model can be used directly to generate translations sets it apart from existing segmenters. SSMT is not a preprocessing step in any sense --- it is single model that learns how to translate and how to segment words, and it can be used to generate translations.
    
We evaluate on English $\rightarrow$ (Xhosa, Zulu, Swati, Finnish, Tswana, Afrikaans). As shown in table \ref{language_details}, these languages span 3 morphological typologies and several levels of data availability, so they provide a varied test suite to evaluate subword methods across different linguistic contexts. 
SSMT outperforms baselines on languages that are agglutinating and conjunctively written (the highest morphological complexity), but is outperformed on simpler morphologies. SSMT achieves its biggest gains on Swati, which is our most data scarce language. 
We conclude that SSMT is justified for morphologically complex languages and especially useful when the languages are low-resourced.

We analyse the linguistic plausibility of SSMT by applying it to unsupervised morphological segmentation. SSMT subwords are closer to morphemes than our baselines.
Lastly, we adapt the methods of \citet{keysers-etal-2020-measuring} to construct an MT test set for morphological compositional generalisation --- the ability to generalise to previously unseen combinations of morphemes.
The performance of all models degrade on the more challenging test set, but SSMT exhibits the greatest robustness. We posit that SSMT's performance gains on morphologically complex languages are due to its morphologically consistent segmentations and its superior modelling of morphological composition.\footnote{Our code and models are available at \url{https://github.com/francois-meyer/ssmt}.}

\section{Related Work}

Subword segmentation has been widely adopted in NLP.
Several algorithms have been proposed, with BPE \citep{sennrich-etal-2016-neural} and ULM \citep{kudo-2018-subword} among the most popular. 
BPE starts with an initial vocabulary of characters and iteratively adds frequently co-occuring subwords.
ULM starts with a large initial vocabulary and iteratively discards subwords based on the unigram language model likelihood. 
%ULM is a probabilistic algorithm that assume independence among subword occurrences (i.e.
%
%like BPE \citep{sennrich-etal-2016-neural} and ULM \citep{kudo-2018-subword} are applied to datasets  
% BPE exemplifies this paradigm. For step 1, it starts with an initial vocabulary of characters and iteratively adds frequently co-occuring vocabulary items to the vocabulary as new subwords. For step 2, it segments words by starting with characters and iteratively merging subwords from the vocabulary (again in order of co-occurrence frequency) until no more merges are posible. 
% ULM also falls in this paradigm, but its training (step 1) and segmentation (step 2) use probabilistic algorithms that assume independence among subword occurrences (i.e. a unigram language model). 
Both of these exemplify the dominant paradigm in NLP:  subword segmentation as a preprocessing step. Segmenters are applied to datasets before models are trained on the segmented text.
%before model training. Models are trained on datasets in which the words have already been segmented into subwords. 

\begin{figure}[t] 
\small
	\centering
	\begin{tabular}{l>{\centering\arraybackslash}p{1.5cm}cc} 
		\toprule
		%&&&&  \multicolumn{3}{c}{\textbf{Training data}}\\
		\textbf{Language} & \textbf{Morphology} & \textbf{Orthography} &  \textbf{Sentences} \\
		\midrule
		Xhosa& \multirow{4}{*}{agglutinative}  & \multirow{4}{*}{conjunctive} & 8.7mil  \\
		Zulu &  & &  3.9mil    \\
            Finnish &  & &  1.6mil  \\
            Swati &  & & 165k  \\
		\midrule
		Tswana & agglutinative & disjunctive &  5.9mil   \\
		\midrule
		Afrikaans & analytic & disjunctive &  1.6mil   \\
		\bottomrule
	\end{tabular}
	\captionof{table}{Morphological typology and training data sizes for the target languages used in our experiments.} 	\label{language_details}
	%\vspace{-2cm}
\end{figure}

There are downsides to relegating subword segmentation to the domain of preprocessing. The algorithms are task-agnostic. BPE is essentially a compression algorithm \citep{gage-1994-bpe}, while ULM assumes independence between subword occurrences. Neither of these strategies are in any way connected to the task for which the subwords will eventually be used (in our case machine translation).
% Such segmenters are also language-agnostic. They are applied without change to different languages, even though different morphologies might require different segmentation techniques. 
%This is because they operate completely independent of model training. Models are reliant on the quality of the subwords produce by segmenters. 
Ideally subword segmentation should be part of the learnable parameters of a model, so that it can be adjusted to optimise the training objective. 

There has been some research on unifying subword segmentation and machine translation. 
Following recent character-based language models \citep{clark-etal-2022-canine, tay2022charformer}, there has been work on character-level NMT models that learn latent subword representations \citep{edman-etal-2022-subword}. However, \citet{libovicky-etal-2022-dont} found that subword NMT models still outperform their character-level counterparts.
\citet{kreutzer-sokolov-2018-learning} learn source sentence segmentation during training and find that models prefer character-level segmentations. DPE \citep{he-etal-2020-dynamic} learns target sentence segmentation during training and is then applied as a subword segmenter.

\begin{figure*} 
%\vspace{-0.5cm}
  \includegraphics[width=\textwidth]{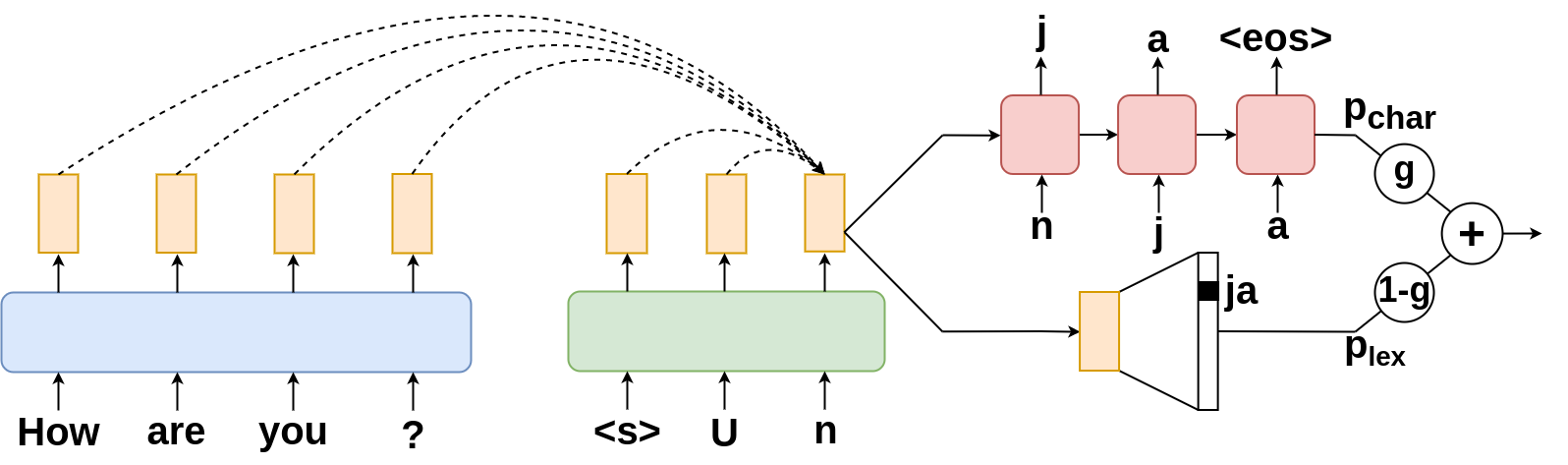}
  \caption{SSMT translates ``How are you?'' to the Zulu ``Unjani?'', computing the probability for subword ``ja''.  
  A Transformer encoder-decoder encodes the BPE-segmented source sentence and character-level target sentence. A mixture between a character decoder and lexicon model (Equation \ref{mixture}) produces the next subword probability.}
  \label{ssmt_architecture}
\end{figure*}

This line of work is related to a more general approach known as segmental sequence modelling, where sequence segmentation is viewed as a latent variable to be marginalised over during training. It was initially proposed for tasks like handwriting recognition \citep{kong-etal-2015-segmental} and speech recognition \citep{wang-etal-2017-sequence}. Subsequently segmental language models (SLMs) have been proposed for unsupervised Chinese word segmentation \citep{sun-deng-2018-unsupervised, kawakami-etal-2019-learning, downey-etal-2021-masked}. 
This was adapted for subword segmentation by \citet{meyer-buys-2022-subword}, who proposed subword segmental language modelling (SSLM). This is the line of work we build on in this paper, adapting subword segmental modelling for NMT. 

Our model contrasts with DPE in a few ways. Firstly, our lexicon consists of the $V$ most frequent character n-grams, so unlike DPE we don't rely on BPE to build the vocabulary. Secondly, we supplement our subword model with a character decoder, which is capable of generating out-of-vocabulary subwords. Lastly, through our proposed dynamic decoding we use SSMT directly to generate translations, instead of having to train an additional NMT model from scratch on our segmentations.

\begin{figure*} 
	%\vspace{-0.5cm}
	\includegraphics[width=\textwidth]{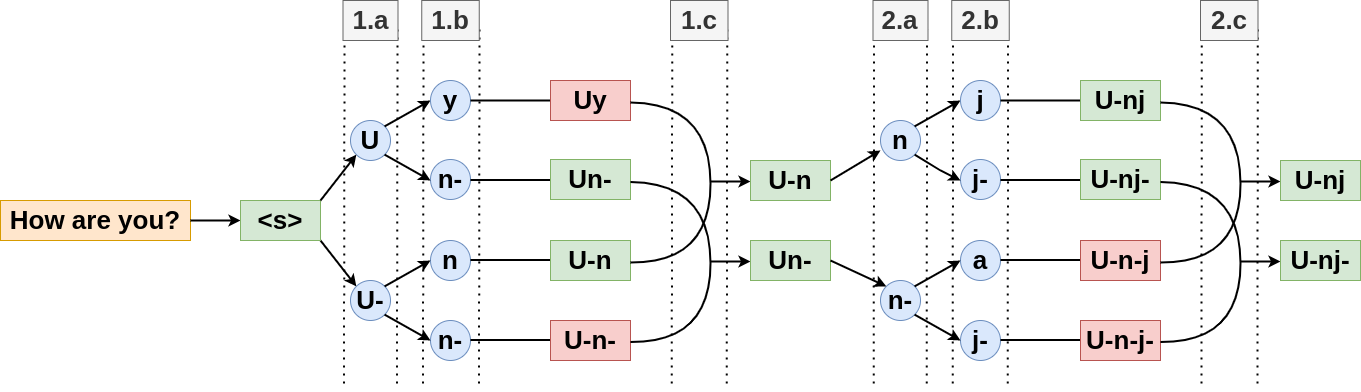}
	\caption{Dynamic decoding for the first 2 characters of a translation (``-'' are subword boundaries). Step (a) produces candidate characters that continue and end the subword. Step (b) peaks one character ahead. Step (c) finalises the segmentation decision. Green sequences are chosen ahead of red ones based on higher sequence probabilities.}
	\label{dynamic_decoding}
\end{figure*}

%They also find that supervised morphological segmenters generally outperform unsupervised segmenters like BPE, especially for agglutinative languages.
% They also find that subword segmentation is particularly beneficial for low-resource languages,
% % because it improves performance on rare and unseen words,
% but on average a simple character n-gram method outperforms BPE \cite{zhu-etal-2019-importance}.

\section{Subword Segmental Machine Translation (SSMT)}

%SSMT learns how to segment target sentence words while training for translation. 
%We model subword segmentation as a latent variable and marginalise over it with a dynamic programming algorithm. 
%To use SSMT during testing, we propose a decoding algorithm that dynamically segments words while generating translations.

%The resulting model can be used to segment words (by computing the maximising segmentation) and to generate translations (with our dynamic decoding algorithm). 

\subsection{Architecture}

SSMT is a Transformer-based encoder-decoder (Figure \ref{ssmt_architecture}). The encoder is that of a vanilla Transformer NMT model. Source language sentences are pre-segmented with BPE. The decoder adapts the subword segmental architecture of \citet{meyer-buys-2022-subword} to be Transformer-based (as opposed to their LSTM-based model) and conditioned on the source sentence. During training SSMT considers all possible subword segmentations of the target sentence and learns which of these optimise its translation training objective.

Given a source sentence of BPE tokens $\mathbf{x} = x_1, x_2, ..., x_{|\mathbf{x}|}$, SSMT generates the target sentence     characters $\mathbf{y} = y_1, y_2, ..., y_{|\mathbf{y}|}$ as a sequence of subwords $\mathbf{s} = s_1, s_2, ..., s_{|\mathbf{s}|}$. We introduce a conditional semi-Markov assumption, whereby each subword probability is computed as
\begin{align} 
    p(s_{i} | \mathbf{s_{< i}, \mathbf{x}}) &\approx p(s_{i} | \pi(\mathbf{s_{< i}}), \mathbf{x}) \\
    & = p(s_{i} | \mathbf{y_{<j}}, \mathbf{x}), \label{segment_markov}
\end{align}
where $\pi(\mathbf{s_{< i}})$ is a concatenation operator that converts the sequence $\mathbf{s_{< i}}$ into the raw unsegmented characters $\mathbf{y_{<j}}$ preceding subword $\mathbf{s_{i}}$. Conditioning on the unsegmented history enables efficiency when we marginalise over subword segmentations. 
%We condition on the source sentence and target sentence history 
%by passing the final output embedding of the decoder to model components.

The subword probability of Equation \ref{segment_markov} is based on a mixture (shown on the right in Figure \ref{ssmt_architecture}),
\begin{align} \label{mixture}
    p(s_{i} | \mathbf{y_{<j}}, \mathbf{x}) =\,\, & g_j p_{\mathrm{char}} (s_{i} | \mathbf{y_{<j}}, \mathbf{x}) + \nonumber\\
    & (1-g_j) p_{\mathrm{lex}} (s_{i} | \mathbf{y_{<j}}, \mathbf{x}),
\end{align}
which combines probabilities from a character LSTM decoder ($p_{\mathrm{char}}$) and a fully connected layer that outputs a probability ($p_{\mathrm{lex}}$) if $s_i$ is in the lexicon. The lexicon contains the $V$ most frequent character sequences ($n$-grams) up to some maximum segment length in the training corpus ($V$ is a prespecified vocabulary size). The lexicon models frequent subwords (e.g. common morphemes), while the character decoder models rare subwords and previously unseen words (e.g. it can copy names from source to target sentences).
The mixture coefficient $g$ (computed by a fully connected layer) allows SSMT to learn, based on context, when the next subword is likely to be in the lexicon and when it should rely on character-level generation.

%The mixture coefficient $g$ is the output of a fully connected layer, so SSMT learns when it can rely on the lexicon (for ) and when it should fall back on the character-level decoder (). 

% FROM SSLM paper: The mixture coefficient gk is also computed from hk with a fully connected neural layer, so the model can learn when to rely on the lexicon and when to revert to character-by- character generation. The

\subsection{Training}

We use this architecture to train a model that jointly learns translation and target-side subword segmentation. The subword segmentation of a target sentence is treated as a latent variable and marginalised over to compute the probability
\begin{align} \label{marginal} 
    p(\mathbf{y} | \mathbf{x}) = \sum_{\mathbf{s}: \pi(\mathbf{s}) = \mathbf{y}} p(\mathbf{s} | \mathbf{x}),
\end{align}
where the probability of a specific subword segmentation $\mathbf{s}$ is computed with the chain rule as a product of its individual subword probabilities (each computed as Equation  \ref{mixture}).

We can compute this marginal efficiently with a dynamic programming algorithm, where at each character position $k$ in the raw target sentence $\mathbf{y}$ the forward probability is,
\begin{align} \label{dp}
    \alpha_k = \sum_{j= f(\mathbf{y}, k)}^{k} \alpha_k p(s = \mathbf{y_{j:t}} | \mathbf{y_{<j}}, \mathbf{x}), 
\end{align}
with $\alpha_0 = 1$. The function $f(\mathbf{y}, k)$ outputs the starting index of the longest possible subword ending at character $k$. This will either be $k - m$, where $m$ is the maximum segment length (a pre-specified hyperparameter) or it will be the starting index of the current word (if character $k-m$ precedes the start of the current word).

% \begin{figure*}[t] 
% 	\centering
% 	\begin{tabular}{lcccc} 
% 		\toprule
% 		%&&&&  \multicolumn{3}{c}{\textbf{Training data}}\\
% 		\textbf{Language} & \textbf{Morphology} & \textbf{Orthography} & \textbf{Family} & \textbf{\# Sentences} \\
% 		\midrule
% 		isiXhosa& \multirow{4}{*}{agglutinative}  & \multirow{4}{*}{conjunctive} & Nguni & 8.7mil  \\
% 		isiZulu &  & & Nguni & 3.9mil    \\
% 		Siswati &  & &Nguni & 165k  \\
%             Finnish &  & & Uralic & 1.6mil  \\
% 		\midrule
% 		Setswana & agglutinative & disjunctive & Sotho–Tswana & 5.9mil   \\
% 		\midrule
% 		Afrikaans & analytic & disjunctive & Germanic & 1.6mil   \\
% 		\bottomrule
% 	\end{tabular}
% 	\captionof{table}{Linguistic information and training data sizes for the languages used in our experimentsgf.} 	\label{language_details}
% 	%\vspace{-2cm}
% \end{figure*}

This last constraint is critical, since it limits the model to learn segmentation of words into subwords. The function $f(\mathbf{y}, k)$ ensures that our model cannot consider segments that cross word boundaries; 
%Our model is learning subword segmentation, as opposed to general sequence segmentation. t
the only valid segments are those within words.
Characters that separate words (e.g. spaces and punctuation) are treated as 1-character segments. In this way we also implicitly model the beginning and end of words, since these are the boundaries of valid segments.

\section{Dynamic decoding} \label{subsec_dynamic_decoding}

For standard subword models, beam search over the subword vocabulary is the \emph{de facto} approach.  
However, the SSMT mixture model (Equation \ref{mixture}) has two vocabularies, a character vocabulary and a subword lexicon. Beam search can be applied to either one. However, to approximate finding the highest scoring translation, subword prediction should be based on the full mixture distribution. 
%To do this our decoding algorithm has to incorporate information from both the character and lexicon models.

During training SSMT considers all possible segmentations of the target sentence with dynamic programming. We would like to consider different segmentations during decoding as well, instead of being limited to the subword boundaries dictated by greedy prediction. Doing this requires retaining part of the dynamic program during decoding, similar to \citet{yu-etal-2016-online} who modelled the latent alignment between (multi-word) segments. 
%and \citet{buys-blunsom-2018-neural}. 
%This enables SSMT to dynamically adjust preferred segmentation as new subwords are generated.  
%We therefore have two requirements for our decoding algorithm: (1) it must incorporate both the character and lexicon models in subword prediction and (2) it must be capable of dynamically adjusting preferred subword segmentations during generation. Designing an algorithm that satisfies these requirements is non-trivial. 
In this section we outline \emph{dynamic decoding}, an algorithm that (1) incorporates both the character and lexicon models and (2) dynamically adjusts subword segmentation during generation. 

% For simplicity, we ex

\subsection{Next character prediction}

Dynamic decoding generates one character at a time and computes next-character probabilities with the full mixture model. Since we generate characters we also explicitly model subword boundary decisions, i.e., when we generate a character we consider whether the character ends a subword (it is the last character in the subword) or whether it continues a subword (more characters will follow in the subword). The mixture model's next-character probability calculation is different, depending on whether we compute the probability of the next character ending the current subword (denoted \textbf{end}) or continuing the current subword  (denoted \textbf{con}). 

Similarly, at each character generation step we have to consider whether the \emph{preceding} character ends or continues a subword. If it ends a subword, then the next character starts a new subword. If the preceding character continues a subword, then the next character is the latest addition to the current subword. These considerations also affects the next-character probability.

Given this setup, we have 4 possible cases for next-character generation: 
\begin{enumerate}[noitemsep,topsep=0pt]
    \item \textbf{con-end} -- the preceding character continues a subword that the next character ends,
    \item \textbf{end-con} -- the preceding character ends a subword and the next character starts a new one,
    \item \textbf{end-end} -- both preceding and next characters end subwords,
    \item \textbf{con-con} -- both preceding and next characters continue the same subword.
\end{enumerate} 
Each case requires different calculations to obtain next-character probabilities with the SSMT mixture model. We present and motivate probability formulas for all 4 cases in Appendix \ref{appendix_nextchar}, defining the  probabilities used in algorithm \ref{dynamic_decoding_alg} ($p_\text{con-end}, p_\text{end-con}, p_\text{end-end}, p_\text{con-con}$). 

% THIS WAS HERE BEFORE
% the preceding character continues a subword that the next character ends (\textbf{con-end}), the preceding character ends a subword and the next character starts and continues a subword (\textbf{end-con}), both preceding and next characters end subwords (\textbf{end-end}), and both preceding and next characters continue the same subword (\textbf{con-con}). 
% Each of these cases require different calculations to obtain next-character probabilities with the SSMT mixture model. We present and motivate probability formulas for all 4 cases in appendix \ref{appendix_nextchar}, defining the  probabilities used in algorithm \ref{dynamic_decoding_alg} ($p_\text{con-end}, p_\text{end-con}, p_\text{end-end}, p_\text{con-con}$). 

\begin{figure*}[t] %\small

	\centering
	\begin{tabular}{lcccccccc} 
		\toprule
		%&&&&  \multicolumn{3}{c}{\textbf{Training data}}\\
		
		Model $\rightarrow$ & \multicolumn{2}{c}{BPE} & \multicolumn{2}{c}{ULM}& \multicolumn{2}{c}{DPE} &  \multicolumn{2}{c}{SSMT}\\
		\cmidrule(){2-9}
		
		English to  $\downarrow$  & BLEU & chrF & BLEU & chrF & BLEU & chrF & BLEU & chrF \\
		\midrule
		Xhosa& 14.3&	\textbf{53.2}	& \underline{\textbf{15.0}} &	\textbf{53.3}& \textbf{14.9}&	\textbf{53.3}	&\underline{\textbf{15.0}} &\underline{\textbf{53.5}}  \\
  
		Zulu& 13.5	&53.2&	\textbf{13.7}&	53.0& \underline{\textbf{14.2}} &	\underline{\textbf{53.7}}	& \underline{\textbf{14.2}}	& \underline{\textbf{53.7}} \\
  
		Finnish& \textbf{15.0}	&\underline{\textbf{50.1}}	& \textbf{15.0}&	49.6& \underline{\textbf{15.4}}	& \textbf{50.0} &	 \textbf{14.4}&	\underline{\textbf{50.1}} \\
  
            Swati&  0.2&	23.4	&0.4	&23.7& 0.3 &	23.5 &	\underline{\textbf{0.7}} &	\underline{\textbf{26.2}} \\
            
		\midrule
		Tswana& \underline{\textbf{10.2}}&	\underline{\textbf{36.9}}&	\textbf{10.1}&	35.5& 9.1&	34.6&	\textbf{9.7}&	36.5 \\
  
		\midrule
		Afrikaans& 33.4&	64.2&	33.5&	64.3&	\underline{\textbf{34.6}}	& \underline{\textbf{65.0}} & 32.0&	63.6 \\

		\bottomrule
	\end{tabular}
	\captionof{table}{MT test set performance (FLORES devtest). \underline{Underline} indicates best BLEU and chrF scores, while \textbf{bold} indicates scores with differences from the best that are not statistically significant ($p$-value of 0.05)} 	\label{mt_results}
	%\vspace{-2cm}
\end{figure*}

\begin{algorithm}[t] 
\small
\SetAlgoLined
    \textbf{Input:} $\mathbf{x}$ is a source sentence of BPE tokens\\
    \textbf{Output:}  $\mathbf{y^*}$ is the generated translation, a character sequence concluding with <eot> (end-of-translation)\\
    \textbf{Notation:}  $C$ is a character vocabulary \\
    $\mathbf{y_\mathrm{\textbf{end}}}$: partial translation, last char ends subword\\
    $\mathbf{y_\mathrm{\textbf{con}}}$: partial translation, last char continues subword \\
    \vspace{0.2cm}
    $y_\mathrm{con} =\underset{y \in C}{\arg\max} \, p_\text{end-con}(y|\mathbf{x}), \mathbf{y_\mathrm{\textbf{con}}} = [y_\mathrm{con}]$ \\
    $y_\mathrm{end} =\underset{y \in C}{\arg\max} \, p_\text{end-end}(y|\mathbf{x}) , \mathbf{y_\mathrm{\textbf{end}}} = [y_\mathrm{end}]$ \\
 
    \vspace{0.1cm}
    \While{$\mathbf{y_\textrm{\textbf{end}}}[-1] \neq $ <eot>}{
        $y_\text{con-con} =\underset{y \in C}{\arg\max} \, p_\text{con-con}(y|\mathbf{y_\mathrm{\textbf{con}}}, \mathbf{x})$ \\
        $y_\text{end-con} =\underset{y \in C}{\arg\max} \, p_\text{end-con}(y|\mathbf{y_\mathrm{\textbf{end}}}, \mathbf{x})$ \\
        $\mathbf{y_\mathrm{\textbf{con}}} = \underset{\mathbf{y} \in \{[\mathbf{y_\mathrm{\textbf{con}}}, y_\text{con-con}], [\mathbf{y_\mathrm{\textbf{end}}}, y_\text{end-con}] \}}{\arg\max \,p(\mathbf{y})}$  \\
        \vspace{0.2cm}
        $y_\text{con-end} =\underset{y \in C}{\arg\max} \, p_\text{con-end}(y | \mathbf{y_\mathrm{\textbf{con}}}, \mathbf{x})$ \\
         $y_\text{end-end} =\underset{y \in C}{\arg\max} \, p_\text{end-end}(y | \mathbf{y_\mathrm{\textbf{end}}}, \mathbf{x})$ \\
        $\mathbf{y_\mathrm{\textbf{end}}} = \underset{\mathbf{y} \in \{[\mathbf{y_\mathrm{\textbf{con}}}, y_\text{con-end}], [\mathbf{y_\mathrm{\textbf{end}}}, y_\text{end-end}] \}}{\arg\max \,p(\mathbf{y})}   $\\
    }
    $\mathbf{y^*} = \mathbf{y_\mathrm{\textbf{end}}}$ \\
    return $\mathbf{y^*}$
 \caption{Dynamic decoding}
 \label{dynamic_decoding_alg}
\end{algorithm}

\subsection{Dynamic segmentation} 

One could use next-character probabilities to greedily generate translations one character at a time, inserting subword boundaries when $p_\text{con-end} > p_\text{con-con}$ or $p_\text{end-end} > p_\text{end-con}$. 
However, this would amount to a greedy search over the space of possible subword segmentations, which might be sub-optimal given characters that are generated later. 
%Once a subword is ended, that subword boundary would be final. Ideally we would like our decoding algorithm to reconsider past segmentation decisions if they turn out to be sub-optimal, given characters that are generated after them. 
A naive beam search would not distinguish between complete and incomplete subwords, which introduces a bias towards short subwords during decoding. 
Ideally the decoding algorithm should make the final segmentation decision based on characters to the left and right of a potential subword boundary, without directly comparing complete and incomplete subwords. 
%This will allow the algorithm to base segmentation decisions on characters to the left and right of a potential subword boundary, instead of inserting boundaries based only on preceding characters. 
To achieve this we design a decoding algorithm that retains part of the dynamic program during generation (see algorithm \ref{dynamic_decoding_alg}). 

%Tby being able to consider different possible segmentations during decoding.
% , greedily or with beam search. Since we have formulas for when the next character continues and ends the current subword, during decoding we can decide whether to end or continue a subword (by comparing the probabilities).

%to consider multiple possible subword segmentations.
% Our decoding algorithm generates one character at a time, while also dynamically deciding when to end the current subword and start a new one. We have shown that different formulas are required to compute probabilities of next characters that continue subwords (equations \ref{next_char1} and \ref{next_char3}) and next characters that conclude subwords (equation \ref{next_char2}). We could simply take the top $k$ candidates based on both of these formulas a. However, this would amount to a greedy search on the dynamic program for subword segmentation. We design an algorithm that retains part of the dynamic program by being able to consider different possible segmentations during decoding.

For simplicity we explain dynamic decoding for a beam size of 1. Figure \ref{dynamic_decoding} demonstrates the generation of the first few characters of a translation. 
The key is to hold out on finalising segmentations until subsequent characters have been generated. 
We compute candidates for the next character, but do so separately for candidates that continue the current subword and those that end the current subword (step (a) in Figure \ref{dynamic_decoding}). The segmentation decision is postponed until after the next character has been generated. We now essentially have two ``potential'' beams --- one for continuing the current subword and another for ending it. For each of these potential beams, we repeat the previous step: we compute candidates for the next character, keeping separate the candidates that continue and end the subword (step (b) in Figure \ref{dynamic_decoding}).

Now we reconsider past segmentations. We compare sequence probabilities across the two potential beams of the character generated one step back (comparisons are visualised by arcs under step (c)). We select the best potential beam that continues the current subword and the best potential beam that ends the current subword. We then repeat the process on these new potential beams.
Essentially we are retrospectively deciding whether the previous character should end a subword. Since we have postponed the decision, we are able to consider how it would affect the generation of the next character. For example, in step (2.c) of Figure \ref{dynamic_decoding}, the subword boundary after character ``n'' is reconsidered and discarded, given that it leads to lower probability sequences when we generate one character ahead.

During training, we consider all possible subword segmentations of a target sentence. During decoding, at each generation step we consider all possible segmentations of the two most recently generated characters. In this way we retain part of the dynamic program for subword segmentation.

% It dynamically adjusts segmentations during generation, allowing the model to insert subword boundaries that better approximate maximising translation probabilities.  

\begin{figure}[t] \small
    \centering
	\begin{tabular}{lc} 
		\toprule
	     \textbf{Model} & \textbf{chrF} \\
        \midrule
        \multicolumn{2}{l}{\textbf{2 models to segment + translate with beam search}} \\      
        \midrule
          +BPEvocab --char (DPE) & 23.3\\
          +lexicon --char (SSMT --char) & 23.7\\
          +lexicon +char (SSMT) & 23.1\\   
          \midrule
         \multicolumn{2}{l}{\textbf{1 model with dynamic decoding}} \\      
        \midrule
	     +lexicon --char (SSMT --char) & 26.2 \\
          +lexicon +char (SSMT) & \textbf{26.4}\\
		\bottomrule
	\end{tabular}
	\captionof{table}{English $\rightarrow$ Swati validation set performance.} 	\label{ablation}
	%\vspace{-2cm}
\end{figure}

\section{Machine Translation Experiments} \label{sect4_mt}

We train MT models from English to 6 languages. As shown in table \ref{language_details}, the chosen languages allow us to compare how effective SSMT is across 3 different morphological typologies - agglutinating conjunctive, agglutinating disjunctive, and analytic. 
%This allows us to compare how effective SSMT is across different morphological typologies. 
Most of the languages are agglutinating conjunctive, since prior work has highlighted the importance of subword techniques for morphologically complex languages \citep{klein-tsarfaty-2020-getting, meyer-buys-2022-subword}. For English to Finnish we train on Europarl\footnote{\url{https://www.statmt.org/europarl/}}, while for the other directions we train on WMT22\_African.\footnote{\url{https://huggingface.co/datasets/allenai/wmt22_african}} The parallel dataset sizes are given in table \ref{language_details}. We use FLORES dev and devtest as validation and test sets, respectively. 

Each probability in the SSMT dynamic program (Equation \ref{dp}) requires a softmax computation, so SSMT takes an order of magnitude (10$\times$) longer to train than pre-segmented models. For example, English to Zulu with BPE trained for 1 day, while SSMT trained for 10 days (both on a single A100 GPU). SSMT training times are comparable to those of the DPE segmentation model. 
%Dynamic decoding introduces additional computational complexity. 
On our test sets it takes on average 15 seconds to translate a single sentence (as opposed to our baselines, which take 0.05 seconds per sentence).
We did experiment with naive beam search over the combined lexicon and character vocabularies of SSMT, but this results in much worse validation performance than dynamic decoding (49.8 vs 53.8 chrF on the English to Zulu validation set; see table \ref{beamsizes} in the Appendix). 
We use a beam size of 5 for beam search with our baselines and for dynamic decoding, since this optimised validation performance (table \ref{beamsizes}). 
Further training and hyperparameter details are provided in Appendix \ref{appendix_training}.

\begin{figure*}[t] %\small
	\centering
	%\begin{tabular}{l |ccc| ccc |ccc| ccc} 
	\begin{tabular}{lccccccccc} 
		\toprule
		& \multicolumn{3}{c}{\textbf{Xhosa}} & \multicolumn{3}{c}{\textbf{Zulu}} & \multicolumn{3}{c}{\textbf{Swati}} \\
		%\midrule 
		\cmidrule(lr){2-4} \cmidrule(lr){5-7} \cmidrule(lr){8-10} 
		\textbf{Model} & P & R & F1 & P & R & F1 & P & R & F1  \\
		\midrule
		\textbf{BPE} 
		&37.16&	25.42&	30.19
            &51.57	&29.62&	37.62
            &19.57	&16.17	&17.71
		
		\\ 
		\textbf{ULM} 
		& \textbf{61.22}&	34.65&	44.25
		& \textbf{63.70}&	31.72&	42.35
		& \textbf{52.48}&	45.26&	48.61
		
		\\
		
		\textbf{DPE} 
		&51.52&	44.24&	47.60
           & 59.66&	41.64&	49.05
           & 16.96&	17.00&	16.98
		\\

            	\textbf{SSMT} 
		& 49.55	&\textbf{72.60}&	\textbf{58.90}
		&52.87&	\textbf{66.41}&	\textbf{58.87}
		& 47.47	&\textbf{61.89}&	\textbf{53.73}   \\	
		\bottomrule
	\end{tabular}
	\captionof{table}{Morpheme boundary identification performance across all words in the morphologically annotated dataset.} 	\label{ums2}
	%\vspace{-2cm}
\end{figure*}

\subsection{MT Results}

We evaluate our models with BLEU and chrF. The chrF score \citep{popovic-2015-chrf} is a character-based metric that is more suitable for morphologically rich languages than token-based metrics like BLEU \citep{google-2022-building}. 
%We tune our models based on validation chrF scores and focus on chrF scores when discussing our test set results. 
MT performance metrics on the full test sets are shown in table \ref{mt_results}. We perform statistical significance testing through paired bootstrap resampling \citep{koehn-2004-statistical}.
In terms of chrF, SSMT outperforms or equals all baselines on all 4 agglutinating conjunctive languages. The same holds for BLEU on 3 of the 4 languages. 

These results prove that SSMT is an effective subword approach for morphologically complex languages. They also corroborate the findings of \citet{meyer-buys-2022-subword} that subword segmental modelling leads to greater consistency across different morphologically complex languages. %DPE also performs well on these languages. 
On Xhosa, Zulu, and Finnish, SSMT and DPE exhibit comparable performance. However, DPE requires multiple training steps: a DPE segmenter model, applying that to a corpus, and then training a NMT model on the segmented corpus. SSMT has the notable benefit of being a single model for segmentation and generation.  

On the languages with simpler morphologies (Tswana and Afrikaans), SSMT is outperformed by  baselines. There is a sharp contrast between the relative performance of SSMT on the morphologically complex and morphologically simple languages. SSMT does not seem to be justified for languages that are not agglutinating and conjunctive. 
%Interestingly, DPE does very well on Afrikaans (an analytic languages), but quite poorly on Tswana (a agglutinating disjunctive languages).  

\subsection{Low-resource translation analysis} \label{siswati_analysis}

SSMT improves performance most drastically on Swati, which is distinct among the translation directions in being extremely data scarce. We confirm that this is not simply because of particular hyperparameter choices, because the finding holds across different settings during hyperparameter tuning (see Figure \ref{ablation} in the Appendix). To investigate the factors behind SSMT's success, we perform an ablation analysis of the different components of SSMT (shown in table \ref{ablation}) compared to DPE.

Learning a subword vocabulary with BPE (the approach of DPE) does not improve performance over the frequency-based lexicon of SSMT.
Our results also show that when the goal is to use the model as a segmenter, supplementing the subword model with a character model worsens performance. 
Dynamic decoding is the most important factor in the success of SSMT. The largest gains do not come from learning subword segmentation during training, but from using the same model directly during inference with dynamic decoding. Having a single model for segmentation, MT, and generation leads to the best performance overall.

%Models trained on so little data can be particularly sensitive to hyperparamer values, so we perform an extensive analysis on Swati validation performance across different settings to test whether SSMT reliably improved performance. The results from this analysis are visualised in figure \ref{ssw_chart}. SSMT consistently outperforms the baselines on Swati translation, even when we compare across different model sizes and subword vocabulary sizes. Given these findings, SSMT seems to hold particular promise for morphologically complex languages that are  extremely low-resource. 

\section{Unsupervised Morphological Segmentation} \label{sec_ums}

Morphemes are the primary linguistic units in agglutinative languages.
%and it has been shown that subwords corresponding more closely to morphemes can improve performance \citep{meyer-buys-2022-subword}.
We can analyse to what extent SSMT subwords resemble morphemes by applying it as a segmenter to the task of unsupervised morphological segmentation. The task is fully unsupervised, since our baselines and SSMT models are tuned to optimise validation MT performance and never have access to morphological annotations (they are trained on raw text). The task amounts to evaluating whether these subword segmenters ``discover'' morphemes as linguistic units.

We evaluate our models on data from the SADiLaR-II project  \citep{GAUSTAD2022107994}. The dataset contains 146 parallel sentences in English and 3 of the agglutinating conjunctive languages for which we train MT models (Xhosa, Zulu, Swati). The dataset provides morphological segmentations for all words in the parallel sentences. We apply the preprocessing scripts of \citet{moeng-etal-2021-canonical} to extract surface segmentations. To apply SSMT as a segmenter we use the Viterbi algorithm to compute the highest scoring subword segmentation of a target sentence given the source sentence. We compare SSMT subwords to the baseline segmenters from our MT experiments. 

Table \ref{ums2} reports precision, recall, and F1 for morpheme boundary identification. SSMT has greater F1 scores than any of the baselines across all 3 languages, indicating that generally SSMT learns subword boundaries that are closer to morphological boundaries. SSMT also has the highest recall for all 3 languages, but lower precision. This show that SSMT sometimes over-segments words, which \citet{meyer-buys-2022-subword} also found to be the case for SSLM. Table \ref{ums1} in the Appendix shows similar results for the same task using morpheme identification as metric. 

\section{Morphological Compositional Generalization} \label{sec6_analysis}

SSMT learns morphological segmentation better than standard segmenters, but is it also learning to compose the meanings of words from their constituent morphemes? 
%words does that mean it can generalise better to words consisting of previously unseen combinations of morphemes?
%but does it also learn morphological composition? That is, does SSMT learn to compose 
To investigate this we design an experiment aimed at testing morphological compositional generalisation. 

Compositional generalisation is the ability to compose novel combinations from known parts \citep{partee1984, FODOR19883}. Recent works have investigated whether neural models are able to achieve such generalisation \citep{Lake2017GeneralizationWS, ijcai2020p708, kim-linzen-2020-cogs}. For example, \citet{keysers-etal-2020-measuring} test whether models can handle novel syntactic combinations of known semantic phrases. 
They construct train/test splits with similar phrase distributions, but divergent syntactic compound distributions. 
%Their procedure  
We adapt their approach to construct a test set with a similar morpheme distribution to the train set, but a divergent word distribution. 
This evaluates whether models can handle novel combinations of known morphemes (previously unseen words consisting of previously seen morphemes). Table \ref{genbench} in the Appendix categorises our experiment according to the generalisation taxonomy of \citet{hupkes2022taxonomy}.

%We design an experiment that can be used to evaluate the compositional generalisation of trained models if their train sets are available. Given train and test sets, our procedure extracts examples from the test set to produce a test \emph{subset} that differs systematically the train set. By controlling the magnitude of the difference between the train set and test subset, we can evaluate our models on test sets that require different levels of morphological compositional generalisation. 
%Unlike some previous works, our approach does not require newly trained models. Instead, we evaluate existing models on newly extracted subsets of the test set. Testing our models on subsets that differ systematically from the dataset on which the models were trained, we can evaluate to what extent our models achieve morphological compositional generalisation.

%One approach in this line of work is to split existing datasets into train and test sets with systematically different distributions \citep{hupkes2022taxonomy}.
%The test set should require compositional generalisation i.e. it should contain compounds that do not occur in the train set (or not in the same relative frequency), that are novel combinations of parts that do occur in the train set. 
%Constructing such datasets can be achieved through synthetic data generation, but this fails to account for the full complexity of natural language. 

\subsection{Compound divergence}

\citet{keysers-etal-2020-measuring} propose compound divergence as a metric to quantify how challenging it is to generalise compositionally from one dataset to another.  
We use it to sample a subset of a test set that requires morphological compositional generalisation from a training set.

To compute morpheme distributions we segment our train and test sets into morphemes with the trained morphological segmenters of \citet{moeng-etal-2021-canonical}. 
Following \citet{keysers-etal-2020-measuring}, we refer to morphemes as \emph{atoms} and words as \emph{compounds}. For a dataset $T$, we compute the distribution of its compounds $F_C(T)$ as the relative word frequencies and the distribution of its atoms $F_A(T)$ as the relative morpheme frequencies.
For a train set $V$ and test set $W$ we compute compound divergence $\mathcal{D}_C(V||W)$ and atom divergence $\mathcal{D}_A(V||W)$, respectively quantifying how different the word and morpheme distributions of the train and test sets are (larger divergence implies greater difference). 
We use the definitions of compound and atom divergence proposed by \citet{keysers-etal-2020-measuring} and include these in Appendix \ref{compgen_appendix}.
We implement a procedure (also outlined in Appendix \ref{compgen_appendix}) for extracting a subset of the test set such that $\mathcal{D}_C$ can be specified and $\mathcal{D}_A$ is held as low as possible, producing a test set that requires models trained on $V$ to generalise to new morphological compositions.

\subsection{Results} 

For this experiment we focus on English $\rightarrow$ Zulu translation.
We extract 2 test subsets of 300 sentences each from Zulu FLORES devtest. For the first subset we specified $\mathcal{D}_C^\mathrm{target} = 0.2$, while for the second  $\mathcal{D}_C^\mathrm{target} = 0.3$. We settled on these values since it was not possible to extract test subsets outside this range with equal atom divergence to the train set (around 0.07 for both). The result is 2 test subsets that require varying degrees of morphological generalisation. The subset with $\mathcal{D}_C = 0.3$ is more challenging than the $\mathcal{D}_C = 0.2$ subset, provided the model is trained on the same train set as ours (English-Zulu WMT22 dataset).

\begin{figure} 
\vspace{-0.5cm}
  \includegraphics[width=1.1\linewidth]{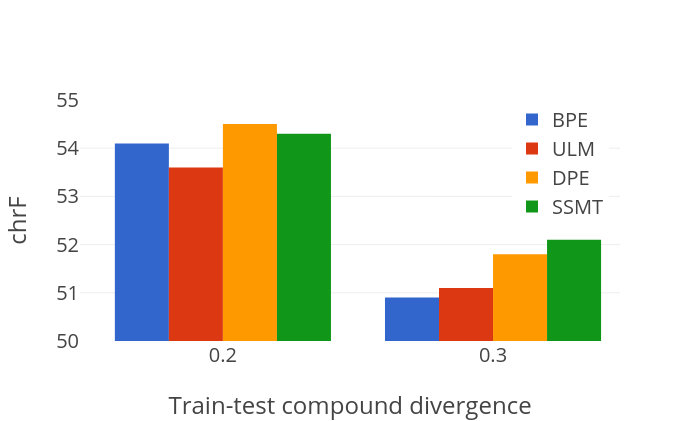}
  \caption{MT performance of our English $\rightarrow$ Zulu models on test subsets that are easier (left) and harder (right) in terms of morphological compositional generalisation.}
  \label{compgen_chart}
\end{figure}

The results are shown in Figure \ref{compgen_chart}. On the less challenging subset ($\mathcal{D}_C = 0.2$), DPE slightly ouperforms SSMT, while the average chrF score of the 4 models is 54.1. On the more challenging subset ($\mathcal{D}_C = 0.3$), the average chrF score drops to 51.5, which shows that models cannot maintain the same level of performance when more morphological generalisation is required. This points to the fact that neural MT models are not reliably learning morphological composition, instead sometimes relying on surface-level heuristics (e.g. learning subword-to-word composition that is not morphologically sound). 
SSMT proves to be most robust to the distributional shift, achieving the best chrF score on the more challenging subset. This shows that SSMT is learning composition more closely resembling true morphological composition. 
%This can be partially attributed to our finding from the previous section that SSMT learns subwords more closely resembling true morphemes. 
SSMT and DPE comfortably outperform BPE and ULM, indicating more generally that learning subword segmentation during training improves morphological compositional generalisation.

\section{Conclusion}

SSMT unifies subword segmentation, MT training, and decoding in a single model. Our results show that it improves translation over existing segmenters when the target language is agglutinative and conjunctively written. It also produces subwords that are closer to morphemes and learns subword-to-word composition that more closely resembles morphological composition. In future work our dynamic decoding algorithm could be used to generate text with subword segmental models for text generation tasks other than MT.

% We believe that this paper opens several avenues for potential future work. Here we put forward 2 possible directions.

% \begin{enumerate}
% 	\item The biggest improvement from SSMT is on the highly data scarce case of English to Siswati translation. SSMT should be tested on more truly low-resource languages, perhaps in combination with other methods for low-resource MT (e.g. multilingual models).
% 	\item We propose dynamic decoding, an algorithm for generating target sentences while adapting subword segmentations one character back. There are many possibilities for alternative decoding algorithms e.g. reconsidering segmentations further into the past or sampling different components of the SSMT model. More generally, dynamic decoding demonstrates the potential of incorporating dynamic programming into a decoding algorithm.
% \end{enumerate} 

\section*{Limitations}

The main downside of SSMT (compared to pre-segmentation models like BPE and ULM) is its computational complexity. Our architecture (Figure \ref{ssmt_architecture}) introduces additional computation in 2 way. Firstly, the decoder conditions on the character-level history of the target sentence, so it has to process more tokens than a standard subword decoder. Secondly, the dynamic programming algorithm (Equation \ref{dp}) requires more computations than standard MT models training on pre-segmented datasets. In practice, SSMT takes an order of magnitude ($10\times$) longer to train than models training on a pre-segmented dataset. Dynamic decoding also adds computational complexity to testing, although this is less of an issue since test set sizes usually permit run times within a few hours.

%For example, our English to Zulu BPE model trained for 1 day, while our SSMT model trained for 10 days (both on a single A100 GPU).  

It would depend on the practitioner to decide whether the performance boosts obtained by SSMT justify the longer training and decoding times.
However, since SSMT is particularly strong for data scarce translation, the computational complexity might be less of an issue. For translation directions like English to Swati, training times are quite short for all models (less than a day for SSMT on subpartitions of the A100 GPU), so the increased training times are manageable.  

% We evaluate our model on languages from a single language group - the Nguni languages. Our findings might not hold for languages with different types of morphological complexity (e.g. fusional languages, where segmentation is difficult because morphemes are fused together). The SSLM achieved consistently good LM performance across all four Nguni languages, but we had to tune the lexicon size and maximum segment length separately for each language. These optimal hyperparameter values varied across languages and would have to be tuned from scratch for new languages.

% The SSLM is able to improve over all baselines as a morphological segmenter, but only if we train it as a word-level sequence model. The long-range SSLM outperforms standard segmenters like BPE and ULM, but performs worse than our entropy-based baselines on F1 scores. This shows that there is a deterioration in segmentation performance when the SSLM is required to model long-range linguistic dependencies - the model tends to over-segment words. 
% We only evaluate our segmentations with automatic evaluation metrics, which provides a rigid, morpheme-based perspective on the segmentation quality. It would be ideal to also include human evaluations of the linguistic plausibility of segmentations. 

% \clearpage

\section*{Acknowledgements}
This work is based on research supported in part by the National Research Foundation of South Africa (Grant Number: 129850).  
Computations were performed using facilities provided by the University of Cape Town’s ICTS High Performance Computing team: \url{hpc.uct.ac.za}.
Francois Meyer is supported by the Hasso Plattner Institute for Digital Engineering, through the HPI Research School at the University of Cape Town. 

% Entries for the entire Anthology, followed by custom entries
\bibliography{anthology,custom}
\bibliographystyle{acl_natbib}

\appendix

\begin{figure*}[t]
	\centering
	%\begin{tabular}{l |ccc| ccc |ccc| ccc} 
	\begin{tabular}{lccccccccc} 
		\toprule
		& \multicolumn{3}{c}{\textbf{Xhosa}} & \multicolumn{3}{c}{\textbf{Zulu}} & \multicolumn{3}{c}{\textbf{Swati}} \\
		%\midrule 
		\cmidrule(lr){2-4} \cmidrule(lr){5-7} \cmidrule(lr){8-10} 
		\textbf{Model} & P & R & F1 & P & R & F1 & P & R & F1  \\
		\midrule
		\textbf{BPE} 
		& 18.04 &	14.23& 	15.91
            & 24.51&	17.52	&20.43
            &9.13	&3.77&	5.33
		
		\\ 
		\textbf{ULM} 
		&\textbf{31.59}&	22.51	&26.29
            &31.47&	20.88&	25.10
            &\textbf{32.31}&	13.72&	19.26
		
		\\
		
		\textbf{DPE} 
		&28.82	&26.16&	27.43
            &33.01	&26.36&	29.31
            &7.97	&3.72&	5.08
		\\

            \textbf{SSMT} 
		& 31.58&	\textbf{41.50}&	\textbf{35.87}
		& \textbf{33.81}&	\textbf{39.57}&	\textbf{36.46}
		& 27.57&	\textbf{15.49}&	\textbf{19.83}
		\\	
		\bottomrule
	\end{tabular}
	\captionof{table}{Morpheme identification performance across all words in the morphologically annotated dataset. Morpheme identification measures how much overlap their is between the subwords in a particular segmentation and the morphemes of a word. } 	\label{ums1}
	%\vspace{-2cm}
\end{figure*}

\section{Next-character probabilities} \label{appendix_nextchar}

Here we present the formulas to compute next-character probabilities with the SSMT mixture model. The probability computations depend on whether the preceding character and next character continue or end subwords, so we provide definitions for all possible subword boundary conditions.
We consider the simplest case first. Given that the previously generated character at position $j-1$ concludes a subword, the probability of the next subword being a single character $y$ is
\begin{align} \label{next_char1} 
    p_\text{end-end}(y | \mathbf{y_{<j}}, \mathbf{x}) =\,\, & g_j p_{\mathrm{char}} (y, \textrm{<eos>}\, |\, \mathbf{y_{<j}}, \mathbf{x}) +  \nonumber \\
     & (1-g_j) p_{\mathrm{lex}}(y | \mathbf{y_{<j}}, \mathbf{x}),
\end{align}
where $\textrm{<eos>}$ is a special end-of-subword token. We can compute this for all $y$ in the character vocabulary and return the top candidates for next character. We modify this for the case where character $j-1$ does not conclude a subword, but character $j$ still does. Then character $j$ constitutes the last character in a subword that started at an earlier character. The probability of next character is then
%The probability of the next character concluding the current subword, which started at character $k$, is
\begin{align} \label{next_char2} 
&\mkern-30mu p_\text{con-end}(y | \mathbf{y_{<j}}, \mathbf{x}) \nonumber \\
&\mkern-30mu =\,\, g_j p_{\mathrm{char}} (y, \textrm{<eos>}\, |\, \mathbf{y_{k:j-1}}, \mathbf{y_{<k}}, \mathbf{x}) +  \nonumber \\
&\,\,\,\, \,\,\,\,\,(1-g_j) p_{\mathrm{lex}}(y | \mathbf{y_{k:j-1}}, \mathbf{y_{<k}}, \mathbf{x}), 
\end{align}
where $k$ is the starting position of the current subword (concluding at $j$) and $\mathbf{y_{k:j-1}}$ are the characters generated so far in the current subword.

These cases still only give us candidates for when the next character concludes a subword. We can modify equation \ref{next_char1} to compute the probability of the next character starting and continuing a subword as
\begin{align} \label{next_char3} 
p_\text{end-con}(y | \mathbf{y_{<j}}, &\mathbf{x}) =\,\,  g_j p_{\mathrm{char}} (y | \mathbf{y_{<j}}, \mathbf{x}) +   \\
& (1-g_j) \sum_{\mathbf{s}: s_1 = y, \mathbf{s} \neq y} p_{\mathrm{lex}}(\mathbf{s} | \mathbf{y_{<j}}, \mathbf{x}). \nonumber
\end{align}
where the first mixture component is simply the probability of the next character under the character-level model (without the $\textrm{<eos>}$ token). The second component marginalises over all subwords starting with $y$. This considers all the possible ways in which the next subword could start with character $y$. It excludes the 1-character subword $y$ ($\mathbf{s} \neq y$), since this constitutes a subword ending with character $j$ (covered by equation \ref{next_char1}). 
Like equation \ref{next_char1}, this covers the case in which the previous character concludes a subword. Similarly to how we generalised equation \ref{next_char1} to equation \ref{next_char2}, we can generalise equation \ref{next_char3} to the case where character $j$ continues a subword started at any given previous character. This produces
\begin{align} \label{next_char4} 
p&_\text{con-con}(y | \mathbf{y_{<j}}, \mathbf{x}) =\,\,  g_j p_{\mathrm{char}} (y |\, \mathbf{y_{k:j-1}}, \mathbf{y_{<k}}, \mathbf{x}) +  \nonumber \\
& (1-g_j) \sum_{\mathbf{s}: s_1 = y, \mathbf{s} \neq y} p_{\mathrm{lex}}(\mathbf{s} |\, \mathbf{y_{k:j-1}}, \mathbf{y_{<k}}, \mathbf{x}). 
\end{align}

% We have now presented formulas that satisfy the first requirement for our decoding algorithm - it incorporates a mixture of the character and lexicon models in computing next-character probabilities. 

\section{Training details} \label{appendix_training}

SSMT is implemented as a sequence-to-sequence model in the fairseq library.
For all our MT models we used the training hyperparameters of the fairseq transformer-base architecture\footnote{\url{https://github.com/facebookresearch/fairseq/blob/main/fairseq/models/transformer/transformer_legacy.py}} (6 encoder layers, 6 decoder layers). 
We extensively tuned the vocabulary sizes of our models on both English-Xhosa and English-Zulu (including separate vocabularies). Validation performance peaked for both at a shared vocabulary of 10k subwords for the baselines. For SSMT it peaked at 5k BPE subwords for the source language and 5k subwords in the target language lexicon. We applied these vocabulary settings to the remaining languages (excluding Swati, which we tuned separately). 

Our SSMT subwords have a maximum segment length of 5 characters, since this was computationally feasible and validation performance did not improve with longer subwords. 
We trained all our models for 25 epochs initially and then continued training until validation performance stopped improving for 5 epochs.  
%We generally found that SSMT requires more epochs of training than our baselines for validation performance to plateau. While baselines often reached peak validation performance around 25 epochs, SSMT kept improving in subsequent epochs.
We trained our DPE segmentation models for 20 epochs (following \citet{he-etal-2020-dynamic}), so DPE required 20 epochs of training for the segmentation model, followed by 25+ epochs for the translation model. We tried sampling ULM segmentations during training for regularisatiion, but initial experiments showed that maximising segmentations led to better validation performance.

\begin{figure}[t] 
%\vspace{-0.5cm}
  \includegraphics[width=\linewidth]{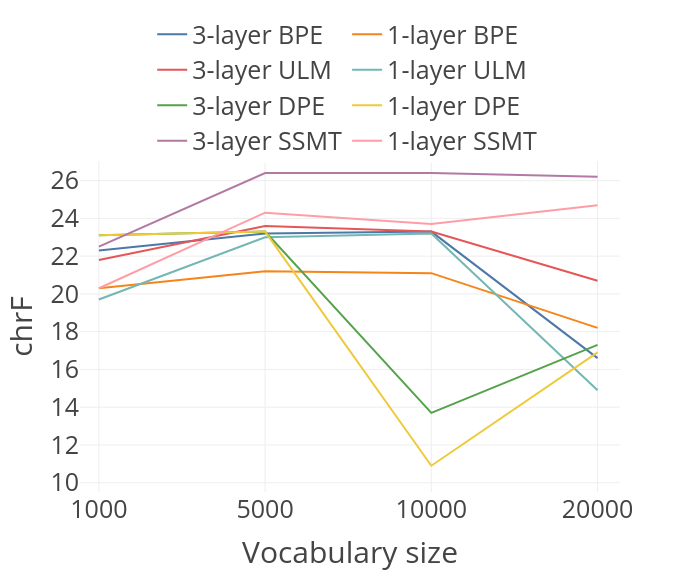}
  \caption{English $\rightarrow$ Swati validation performance.}
  \label{ssw_chart}
\end{figure}

Since models are more sensitive to hyperparameter settings in the data scarce setting \citep{araabi-monz-2020-optimizing}, we performed more extensive hyperparameter tuning for the extremely low-resource case of English $\rightarrow$ Swati. We tuned the number of layers and the vocabulary size (see Figure \ref{ssw_chart}). We found that smaller models (less layers) greatly improved validation performance for all models.

\begin{figure}[t] \small
    \centering
	\begin{tabular}{lcccc} 
		\toprule
            & \multicolumn{2}{c}{Mixture beam search} & \multicolumn{2}{c}{Dynamic decoding} \\
           \cmidrule(lr){2-3}  \cmidrule(lr){4-5}
	     Beam size & BLEU & chrF & BLEU & chrF \\
        \midrule
        
	     1 & 11.8 & 49.7   & 13.6 & 52.2 \\
          3 & 11.5 &  49.2   & 14.1 & 53.6\\
          5 & 11.2 & 49.4   & \textbf{14.5} & \textbf{53.8}\\
          7  & 11.4 & 49.6 & 14.3 & \textbf{53.8}\\
          10 & 11.5 & 49.8 & 14.4 & \textbf{53.8}\\
		\bottomrule
	\end{tabular}
	\captionof{table}{English $\rightarrow$ Zulu validation set performance of SSMT with dynamic decoding compared to standard beam search over the lexicon and character distributions of the SSMT mixture model. Applying standard beam search to SSMT results in poor performance, which justifies the introduction and added computational complexity of dynamic decoding.} 	\label{beamsizes}
	%\vspace{-2cm}
\end{figure}

% Set tabular size
%\newcommand{\tabularwidth}{\textwidth}

% Set symbols for experiments
\newcommand{\expone}{\Checkedbox}
        
% Create table         
\renewcommand{\arraystretch}{1.1}         
\setlength{\tabcolsep}{0mm}         
\begin{figure*}[h]

% Set tabular size
\newcommand{\tabularwidth}{\textwidth}
        
% Create table         
\renewcommand{\arraystretch}{1.1}         
\setlength{\tabcolsep}{0mm}         
\begin{tabular}{|p{\tabularwidth}<{\centering}|}         
\hline
               
% Record the experiments' motivations               
\rowcolor{gray!60}               
\textbf{Motivation} \\               
\footnotesize
\begin{tabular}{p{0.25\tabularwidth}<{\centering} p{0.25\tabularwidth}<{\centering} p{0.25\tabularwidth}<{\centering} p{0.25\tabularwidth}<{\centering}}                        
\textit{Practical} & \textit{Cognitive} & \textit{Intrinsic} & \textit{Fairness}\\
\expone\hspace{0.8mm}	& 	% practical
 & 		% cognitive
\expone\hspace{0.8mm} & 		% intrinsic
 \\		% fairness_inclusivity
\end{tabular}\\
               
% Record the experiments' generalisation type               
\rowcolor{gray!60}               
\textbf{Generalisation type} \\               
\footnotesize
\begin{tabular}{m{0.21\tabularwidth}<{\centering} m{0.2\tabularwidth}<{\centering} m{0.13\tabularwidth}<{\centering} m{0.13\tabularwidth}<{\centering} m{0.13\tabularwidth}<{\centering} m{0.2\tabularwidth}<{\centering}}                   
\textit{Compositional} & \textit{Structural} & \textit{Cross-task} & \textit{Cross-language} & \textit{Cross-domain} & \textit{Robustness}\\
\expone\hspace{0.8mm}& 		% compositional
\expone\hspace{0.8mm}& 		% structural
& 		% across_task
& 		% across_language
& 		% across_domain
 \\		% robustness

\end{tabular}\\
             
% Record the experiments' shift type             
\rowcolor{gray!60}             
\textbf{Shift type} \\             
\footnotesize
\begin{tabular}{p{0.25\tabularwidth}<{\centering} p{0.25\tabularwidth}<{\centering} p{0.25\tabularwidth}<{\centering} p{0.25\tabularwidth}<{\centering}}                        
\textit{Covariate} & \textit{Label} & \textit{Full} & \textit{Assumed}\\  
\expone\hspace{0.8mm}	& 	% covariate
& 		% label
& 		% full
 	\\	% assumed

\end{tabular}\\
             
% Record the experiments' shift source             
\rowcolor{gray!60}             
\textbf{Shift source} \\             
\footnotesize
\begin{tabular}{p{0.25\tabularwidth}<{\centering} p{0.25\tabularwidth}<{\centering} p{0.25\tabularwidth}<{\centering} p{0.25\tabularwidth}<{\centering}}                          
\textit{Naturally occuring} & \textit{Partitioned natural} & \textit{Generated shift} & \textit{Fully generated}\\
& 		% naturally_occurring
\expone\hspace{0.8mm}	 & 	% partitioned_natural_data
& 		% generated_shifts
 	\\	% fully_generated_data

\end{tabular}\\
             
% Record the experiments' shift locus             
\rowcolor{gray!60}             
\textbf{Shift locus}\\             
\footnotesize
\begin{tabular}{p{0.25\tabularwidth}<{\centering} p{0.25\tabularwidth}<{\centering} p{0.25\tabularwidth}<{\centering} p{0.25\tabularwidth}<{\centering}}                         
\textit{Train--test} & \textit{Finetune train--test} & \textit{Pretrain--train} & \textit{Pretrain--test}\\
\expone\hspace{0.8mm}	 & 	% train-test
& 		% finetune_train-test
& 		% pretrain-train
 	\\	% pretrain-test
\end{tabular}\\

\hline
\end{tabular}

\captionof{table}{GenBench evaluation card (\url{https://genbench.org/}) categorising our morphological compositionl generalisation experiment according to the generalisation taxonomy of \citet{hupkes2022taxonomy}.} 	\label{genbench}
	%\vspace{-2cm}
\end{figure*}

\section{Morphological compositional generalisation test subset extraction} \label{compgen_appendix}

For a train set $V$ and test set $W$ we compute the compound divergence and atom divergence, respectively as 
\begin{align} \label{divergence} 
\mathcal{D}_C(V||W) = 1 - C_{0.1}(F_C(V) || F_C(W)), \\
\mathcal{D}_A(V||W) = 1 - C_{0.5}(F_A(V) || F_A(W)),
\end{align} 
where $C_{\alpha}(P||Q)$ is the Chernoff coefficient \citep{CHUNG1989280}. This is a measure of the similarity of 2 distributions $P$ and $Q$ computed as
\begin{align} \label{chernoff} 
C_{\alpha}(P||Q) = \sum_k p_k^\alpha q_k^{1-\alpha},
\end{align}
where $\alpha$ is a parameter that weights the importance of the distributions in the similarity metric. We follow \citet{keysers-etal-2020-measuring} in setting $\alpha = 0.1$ for compound divergence (more important to measure whether or not compounds occur in train than to measure how close the distributions are) and $\alpha = 0.5$ for atom divergence (atom distributions should match as far as possible).

We implement a procedure that, given a train set $V$, extracts a prespecified number of sentences from a test set $W$, such that $\mathcal{D}_C(V||W) = \mathcal{D}_C^\mathrm{target}$ (where $ \mathcal{D}_C^\mathrm{target}$ is the desired compound divergence) and $\mathcal{D}_A(V||W)$ is held as low as possible. 
The procedure starts with the empty test subset and iteratively adds one sentence from the test set. At each step, it randomly samples $k$ sentences from the test set (we set $k=100$) and adds the sentence that minimises
\begin{align} \label{minimises} 
|\mathcal{D}_C - \mathcal{D}_C^\mathrm{target}| + \mathcal{D}_A,
\end{align}
where $\mathcal{D}_C^\mathrm{target}$ is the prespecified compound divergence target for the experiment. Iteratively adding sentences that minimise equation \ref{minimises} results in a test subset containing atoms (morphemes) that the model was exposed to during training, but compounds (words) that it was not. 
We can control the degree of compositional novelty in the test subset compounds by setting  $\mathcal{D}_C^\mathrm{target}$ in our procedure.

\end{document}